\pdfoutput=1
\documentclass{article}
\usepackage{spconf,amsmath,graphicx,parskip}
\usepackage{sectsty}
\usepackage{bbm}
\usepackage{textcomp}
\usepackage{siunitx}
\usepackage{fancyhdr}
\newcommand\numberthis{\addtocounter{equation}{1}\tag{Eq. \theequation}}
\newcommand\norm[1]{\left\lVert#1\right\rVert}
\newcommand{\squeezeup}{\vspace{-2.5mm}}
\usepackage[font=small,labelfont=bf]{caption}
\newcommand\thefontsize[1]{{#1 The current font size is: \f@size pt\par}}

\usepackage{fancyhdr}

\title{Generative Spatiotemporal Modeling Of Neutrophil Behavior}
 \name{Narita Pandhe$^{\star}$ \qquad Balazs Rada$^{\dagger}$ \qquad Shannon Quinn$^{\star}$}
 \address{$^{\star}$ Department of Computer Science \\
     $^{\dagger}$ Department of Infectious Diseases  \\
     University of Georgia \\
     naritapandhe@uga.edu, radab@uga.edu, squinn@cs.uga.edu}
\begin{document}
%
\maketitle
\begin{abstract}
\small{
Cell motion and appearance have a strong correlation with cell cycle and disease progression. Many contemporary efforts in machine learning utilize spatio-temporal models to predict a cell\textquotesingle s physical state and, consequently, the advancement of disease. Alternatively, generative models learn the underlying distribution of the data, creating holistic representations that can be used in learning. In this work, we propose an aggregate model that combine Generative Adversarial Networks (GANs) and Autoregressive (AR) models to predict cell motion and appearance in human neutrophils imaged by differential interference contrast (DIC) microscopy. We bifurcate the task of learning cell statistics by leveraging GANs for the spatial component and AR models for the temporal component. The aggregate model learned results offer a promising computational environment for studying changes in organellar shape, quantity, and spatial distribution over large sequences.
}
\end{abstract}
\vspace{-0.2cm}
\vspace{-6pt}
\setlength{\parskip}{0em}
\squeezeup
\section{Introduction}
\label{sec:intro}
\squeezeup
\vspace{-2.0mm}
Polymorphonuclear neutrophil granulocytes (neutrophils) are the most abundant white blood cells in most mammals. They are highly motile phagocytic cells that constitute the first line of defense of the innate immune system \cite{howneutrophilskill}. Study of neutrophils and their underlying motion patterns provide insights into a host\textquotesingle s response and behavior as a function of specific stimulus. Our understanding of cell behavior and the sources of cellular variation can be significantly aided and tested using cell modeling and simulations \cite{punctuateproteinpattern}. \par
Recently, generative models have been extensively utilized for natural images. Examples include Variational Autoencoders \cite{journals/corr/KingmaW13} and Generative Adversarial Networks (GANs) \cite{NIPS2014_5423}. Generative models have the ability to learn the underlying statistical distributions over data and, thus, can generate exemplars of the true data set. It can learn sophisticated conditional relationships  as well. In 2014, \cite{NIPS2014_5423} proposed Generative Adversarial Networks (GANs), a framework for learning generative models. GANs do not rely on training objectives related to log-likelihood. Instead, GAN training can be seen as a competitive game between two models: the generator ($G$) and the discriminator ($D$). Deep Convolutional GANs (DCGANs) \cite{DBLP:journals/corr/RadfordMC15} train convolutional networks in adversarial settings in order to generate natural images from CelebA \cite{liu2015faceattributes}, LSUN \cite{DBLP:journals/corr/YuZSSX15}, and Imagenet datasets \cite{imagenet_cvpr09}. \cite{osokin2017biogans} applied GANs to biological images to study the coexistence of proteins. Initial GAN models suffered from issues including training instabilities and mode collapse, making them harder to use. Active areas of research include novel applications, optimizing the network architecture, developing best training practices, and improving the cost function. \par
For computer vision systems, motion synthesis is still a challenging task and is drawing more contemporary research attention. Synthesis can be defined as generating new versions of a dataset which follow the distribution of original and is closely related to modeling. \cite{motionfromannotations} presents an algorithm that synthesizes motions based on annotations that describe it. Motion is constructed by splitting segments of movement from a corpus of motion data and assembling them. Each segment is modeled using an autoregressive process. This helps in modeling complicated non-stationary sequences which a single autoregressive process cannot handle. In \cite{armotionsynthesis}, frames of original videos are projected into low-dimensional space and then learned as an AR model. \cite{ad04bba6acd84038b8d6e913b5669b55} extends this approach by overcoming the problems of non-linearities in the data either using a spline-fitting approach or a combined appearance model. Many approaches have employed GANs for video generation and frame prediction. \cite{DBLP:journals/corr/VillegasYHLL17} utilizes two convolutional networks, separating foreground and background imagery, to learn directly from a massive dataset of real-world videos. \par
In this work, we simulate the behavior of human neutrophils. Considering the limited dataset available, we propose an aggregate application of GANs and AR models. We bifurcate our approach as two tasks: generating neutrophil\textquotesingle s appearance and its motion, capturing the statistics independently. The GAN learns the appearance and spatial statistics, while the AR model captures the temporal aspect. Simulation is then achieved by sampling a point from the appearance space given the temporal dynamics up to the last observation. \par
\vspace{-0.58cm}
\section{Related Works}
\squeezeup
\vspace{-1.5mm}
Several computational methods have been proposed for constructing from image data, statistical models of cellular and subcellular structures. General shape models such as Active Shape models \cite{Cootes:1995:ASM:206543.206547} and cell shape model conditional on the nucleus shape \cite{zhao2007automated} have been used. To our knowledge, the closest related literature is comprised of \cite{osokin2017biogans} for biological image synthesis and \cite{DBLP:journals/corr/VillegasYHLL17} for motion synthesis. Differences to \cite{osokin2017biogans} include the following. (1) Our GAN architecture is based on DCGANs, while theirs is a modified DCGAN for channel separation. (2) They apply GANs to samples from fluorescent microscopic images consisting of two channels, red and green. We use GANs for DIC microscopy images consisting of a single channel. They tackle a more difficult problem: using the information contained in the red channel learn how to generate a cell with several green-labeled proteins together. We are modeling single channel cell images. \par
Like this work, \cite{DBLP:journals/corr/VillegasYHLL17} uses a two stream model, but differs as follows. (1) They use Long Short Term Memory Networks (LSTMs), while we use DCGANs for content and derive AR processes from motion. (2) Their network learns the temporal dynamics directly from raw pixels, using identified features combined with spatial features to make pixel-level predictions. We assume the background is stationary and only the foreground cells move. All the pixels of the foreground cells move similarly, so we pool them together into an AR model.\par
\vspace{-0.2cm}
\begin{figure} [htb]
\begin{minipage}[b]{0.48\linewidth}
  \centering
  \centerline{\includegraphics[width=3.8cm]{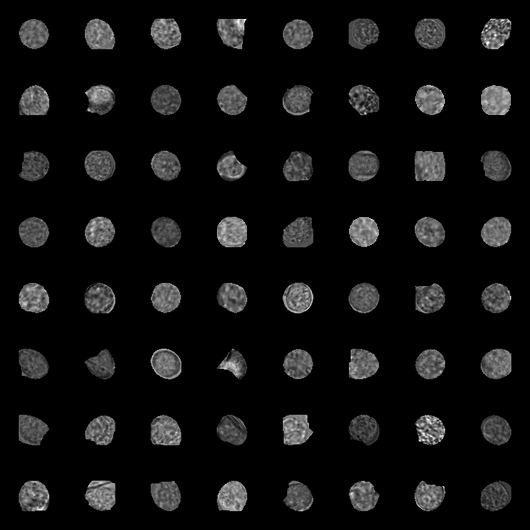}}
  \vspace{-0.3cm}
\end{minipage}
\hfill
\begin{minipage}[b]{0.48\linewidth}
  \centering
  \centerline{\includegraphics[width=3.8cm]{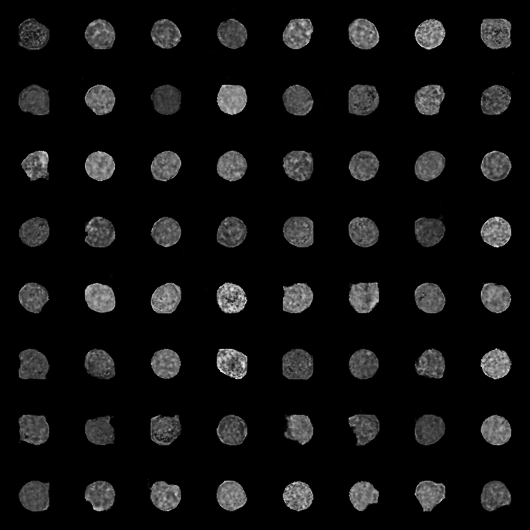}}
  \vspace{-0.3cm}
\end{minipage}
\caption{Real (left) and synthesized (right) images of neutrophil. The synthetic images were created using DCGAN combined with Improved WGAN loss function.
}
\label{fig:res2}
\end{figure}
\squeezeup
\squeezeup
\vspace{-5.5mm}
\section{Data}
\label{sec:data}
\squeezeup
\vspace{-2.5mm}
Videos imaging the two dimensional motion of human neutrophilic granulocytes are provided by Balazs Rada (Department of Infectious Diseases, University of Georgia). The videos are recorded using DIC microscopy which is used to
enhance the contrast in unstained, transparent samples. The dataset consists of 11 videos, including 3 videos of normal neutrophils and 8 videos of neutrophils treated with an inhibitor, MRS2578, targeting a purinergic receptor. Duration
of most of the videos is 3.0secs. We extracted frames of 1024x1024 resolution at 20fps. Individual cells were segmented using fully convolutional DenseNets \cite{DBLP:journals/corr/JegouDVRB16}, centered and resized to 64x64 resolution, resulting in 17280 total grayscale images. \par
\squeezeup
\squeezeup
\vspace{-2.5mm}
\section{Methods}
\label{sec:methods}
\squeezeup
\subsection{GANs for Cell Image Synthesis}
\label{ssec:submethod1}
\squeezeup
\squeezeup
\begin{multline*}
\min_{G} \max_{D} V(D,G) = 
\underset{x \sim p_{data(x)}}{\mathbbm{E}}
\big[\log D(x)\big]
\\ 
+\underset{z \sim p_{z(z)}}{\mathbbm{E}}
\big[\log (1-D(G(z)))\big] \numberthis \label{eqn1}
\end{multline*}
\vspace{-10.5mm}
\begin{multline*}
L = \underbrace{\underset{\widetilde{x} \sim \mathbbm{P}_{g}}{\mathbbm{E}}\big[D(\widetilde{x})\big] - \underset{x \sim \mathbbm{P}_{r}}{\mathbbm{E}}\big[D(x)\big]}
_\text{Original Critic Loss} 
\\
+ \underbrace{\lambda \, \underset{\hat{x} \sim \mathbbm{P}_{\hat{x}}}{\mathbbm{E}}
\big[ (\norm{\nabla_{\hat{x}} D(\hat{x})}_{2}-1)^2\big]}_\text{Gradient Penalty} \numberthis \label{eqn2}
\vspace{-4.0mm}
\end{multline*}
GANs consists of 2 neural networks competing against each other: Generator ($G$) and Discriminator ($D$). $G$ generates images from random noise. While doing so, it tries to get as close as it can, to the distribution of real images. $D$ classifies between the real images and fake images generated by $G$. Both are trying to perform best at their respective tasks and maximise their gains. $D$ is characterized as adversarial loss for training $G$. \par
Formally, consider a set of training images, \( x \in X_{real} \) coming from a real distribution \(P_{d}\). The generator is a neural network \(G (z,\theta G) \) parametrized by \(\theta G\) and discriminator is a neural network \(D(x,\theta D)\) parametrized by \(\theta D\). \(G (z,\theta G) \) takes in random noise \(z\) from the distribution \(P_{z}\) and generates images \( x \in X_{fake} \). \(D(x,\theta D)\) takes images from \( x \in X_{real} \) and \( x \in X_{fake} \), both, and outputs a scalar between [0, 1]. The output is higher if the sample belongs to \(X_{real}\) else \(X_{fake}\). Both $G$ and $D$ are trained simultaneously. The goal of $D$ is to maximize the probability of assigning correct labels to an input while $G$ minimizes \( \log ⁡(1-D(G(z)))\). As a result $D$ and $G$ can be seen as playing a minimax game, as formulated in \ref{eqn1}. Historical attempts to scale up GANs using CNNs to model images have been generally unsuccessful \cite{NIPS2014_5423}. DCGANs \cite{DBLP:journals/corr/RadfordMC15} identified a family of architectures that resulted in stable training and can generate higher resolution images. We have adopted the architecture of DCGAN for both generator and discriminator.\par
\ref{eqn1} can be reformulated via minimization of the Jensen Shannon (JS) divergence between the data-generating distribution \(P_{d}\) and the distribution \(P_{g}\) induced by \(P_{z}\) and $G$. \cite{2017arXiv170107875A} theoretically justified that JS minimized by GANs behaves badly and is potentially not continuous w.r.t to the generator’s parameters. They propose using an alternative distance - Earth Mover\textquotesingle s distance (EM) also known as Wasserstein Distance, W(q,p). Since, computing Wasserstein distance is intractable, \cite{2017arXiv170107875A} shows an approximate solution to the same using Kantorovich-Rubinstein duality, wherein $D$ is the set of 1-Lipschitz functions. To enforce the Lipschitz constraint authors propose to clip the weights of the critic ($D$ referred as critic because it\textquotesingle s not trained to classify) within a compact space $[-c, c]$. Recently, \cite{DBLP:journals/corr/GulrajaniAADC17} proposed an alternative way to enforce the Lipschitz constraint. Instead of weight clipping, they penalize the norm of the critic\textquotesingle s gradient with respect to its input, for random samples \(\hat{x} \sim \mathbbm{P}_{\hat{x}}\). The objective function \ref{eqn2} leads stable training of a wide variety of GAN architectures with almost no hyperparameter tuning. \par
\squeezeup
\squeezeup
\subsubsection{Experiments}
\label{sssec:gansubsubhead1}
\squeezeup
We evaluated performances of models based on DCGAN
architecture trained with GAN, Wasserstein GAN (WGAN), and Improved WGAN loss functions \cite{NIPS2014_5423,2017arXiv170107875A,DBLP:journals/corr/GulrajaniAADC17}. To evaluate the performance of GANs we utilize the optimization-based approach discussed by \cite{osokin2017biogans} to check if the test samples can be reconstructed well. To test for mode collapse, a common failure in GANs, for a fixed trained generator $G$ we examine how well it can reconstruct images from a held out test set. For each image in the test set, we minimize the L2-distance between the generated and test images w.r.t. the noise vector $z$. We use 50 iterations of L-BFGS and select the best reconstruction out of 3 runs. We also report the negative log likelihood (NLL) w.r.t. the prior \(P_{z}\) of the noise vectors $z$. 
\squeezeup
\begin{figure}[htb]
\begin{minipage}[b]{1.0\linewidth}
  \centering
  \centerline{\includegraphics[width=5cm]{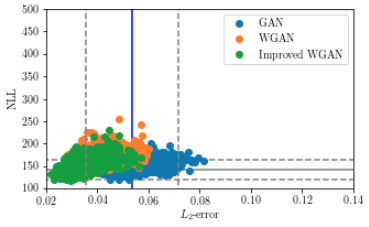}}
  \vspace{-0.3cm}
\end{minipage}
\caption{Reconstruction errors against negative log likelihood (NLL) of the latent vectors found by reconstruction are displayed. The vertical blue line shows the mean L2-error. Horizontal gray line show mean NLL (\(\pm 3\)std) of the noise sampled from the Gaussian prior. Lower values for both are better.}
\vspace{-0.5cm}
\end{figure}
\squeezeup
\vspace{-0.25cm} 
\subsubsection{Latent Space Walk}
\label{sec:gansubsubhead2}
\squeezeup
We can interpolate between points in the latent space and understand the landscape. Walking the manifold can identify if there are any sharp transitions and whether the network has memorized. If walking the latent space results in smooth semantic changes to the image generations we can reason that the model has learned relevant, interesting representations \cite{DBLP:journals/corr/RadfordMC15}.
\squeezeup
\squeezeup
\begin{figure}[htb]
\vspace{-0.2cm}
\begin{minipage}[b]{1.0\linewidth}
  \centering
  \centerline{\includegraphics[width=3cm]{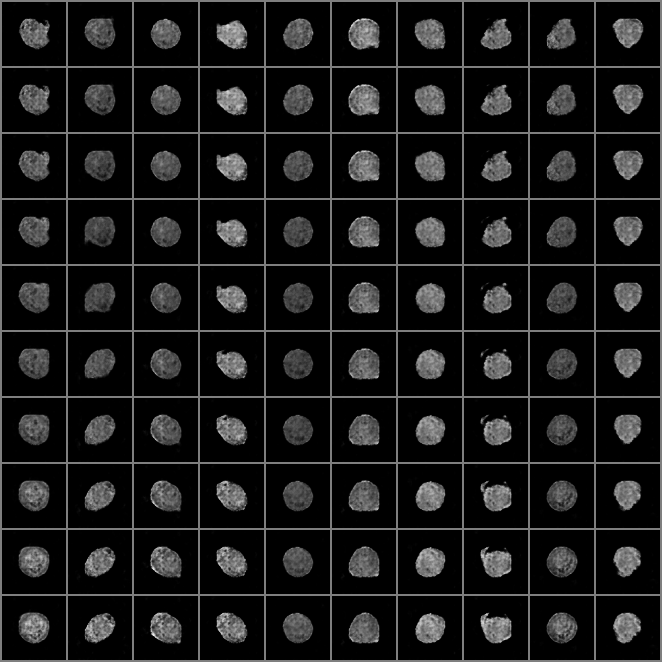}}
  \vspace{-0.3cm}
\end{minipage}
\caption{Interpolation between a series of 10 random points in the latent space depicts
that the space learned has smooth transitions. Top row depicts the starting location for each
of the 10 points. Last row depicts the respective ending locations.}
\label{fig:res3}
\vspace{-0.5cm}
\end{figure}
\squeezeup
\vspace{-0.25cm}
\subsection{AR for Cell Motion Synthesis}
\squeezeup
\squeezeup
\begin{figure} [htb]
\begin{minipage}[b]{.48\linewidth}
  \centering
  \centerline{\includegraphics[width=4.5cm]{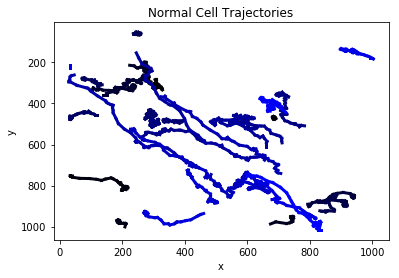}}
  \vspace{-0.3cm}
\end{minipage}
\hfill
\begin{minipage}[b]{0.48\linewidth}
  \centering
  \centerline{\includegraphics[width=4.5cm]{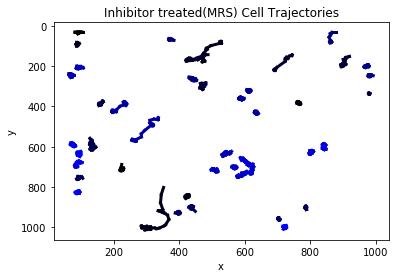}}
  \vspace{-0.3cm}
\end{minipage}
\caption{2D trajectory plots of normal neutrophil and inhibitor-treated (MRS) neutrophil. The inhibitor-treated (MRS) neutrophil tend to exhibit less movements in comparison to the normal ones.}
\vspace{-0.5cm}
\label{fig:res4}
\end{figure}
\squeezeup
\begin{equation*} 
\overrightarrow{y_{t}} = C\overrightarrow{x_{t}} + \overrightarrow{u_{t}} \numberthis \label{eqn3}
\end{equation*}
\begin{equation*} 
\overrightarrow{x_{t}} = B_{1\overrightarrow{x}_{t-1}} + B_{2\overrightarrow{x}_{t-2}}+..+B_{d\overrightarrow{x}_{t-d}}+\overrightarrow{v_{t}} \numberthis \label{eqn4}
\end{equation*}
\label{ssec:submethod2}
\vspace{-0.5mm}
Different motion patterns are observed based on the cell conditions. We build a global motion pattern for normal and inhibited cells respectively, because we assume all the pixels(under the same conditions) move similarly. Based on the existing motion characteristics, new sequences can be synthesized for the corresponding cells. AR models are linear dynamical systems and are able to model a pattern of points in a particular space having a temporal component. 

An AR process for a series of points in a $d$-dimensional space can be modelled as \ref{eqn3}, \ref{eqn4}. \ref{eqn3} decomposes each video frame \(\overrightarrow{y_{t}}\) into a low-dimensional state vector \(\overrightarrow{x_{t}}\) and a white noise term \(\overrightarrow{u_{t}}\). \ref{eqn4} denotes new state \(\overrightarrow{x_{t}}\) is a function of the sum of $d$ of its previous states \(\overrightarrow{x}_{t-1},\overrightarrow{x}_{t-2},..,\overrightarrow{x}_{t-d}\), each multiplied by corresponding coefficients \( B = {B_{1} ,B_{2} ,...,B_{d}}\) \cite{spquinn}. The noise terms $u$ and $v$ represent the residual difference between the observed data and the solutions to the linear equations, assumed to be Gaussian White noise.
\squeezeup
\begin{figure}[htb]
\begin{minipage}[b]{1.0\linewidth}
  \centering
  \centerline{\includegraphics[width=9cm]{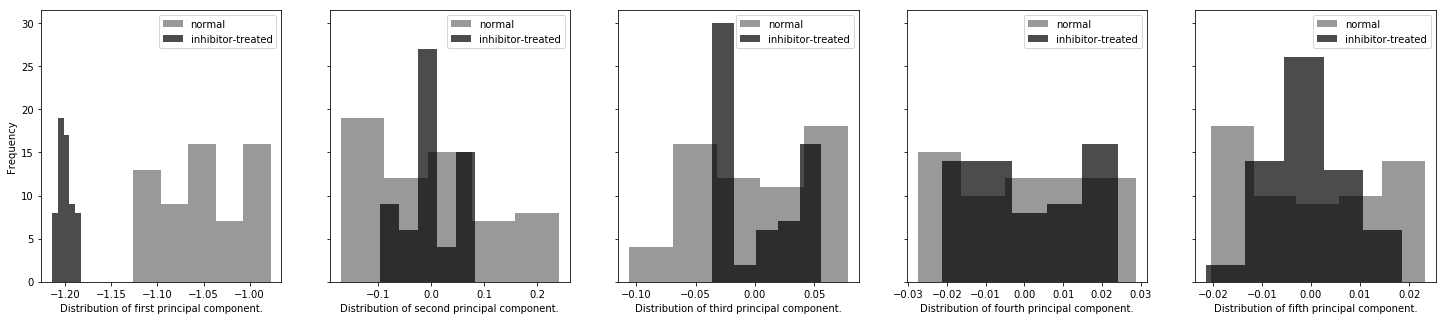}}
  \vspace{-0.3cm}
\end{minipage}
\caption{Histograms show the distributions of values taken by normal (gray) and inhibitor-treated (black) neutrophil for top 5 principal components.}
\vspace{-0.5cm}
\end{figure}
\squeezeup
\vspace{-0.25cm}
\subsubsection{Experiments}
\label{sec:arsubsubhead1}
\squeezeup
Neutrophil motion is represented as trajectories of individual cells consisting of its center Cartesian coordinates across all the frames. Trajectories belonging to normal and inhibited cells are pooled separately and then projected into an eigenspace using SVD, yielding the principal components $C$. Subsequently, AR coefficients are determined. Parameter $q$ determines the dimensionality of the subspace $C$; parameter $d$ determines the order of AR coefficients \(B = {B_{1} ,B_{2} ,...,B_{d}}\). We performed grid search over \(q \in [2,10]\) and \(d \in [1,10]\).
\squeezeup
\begin{figure} [htb]
\begin{minipage}[b]{.48\linewidth}
  \centering
  \centerline{\includegraphics[width=4.0cm]{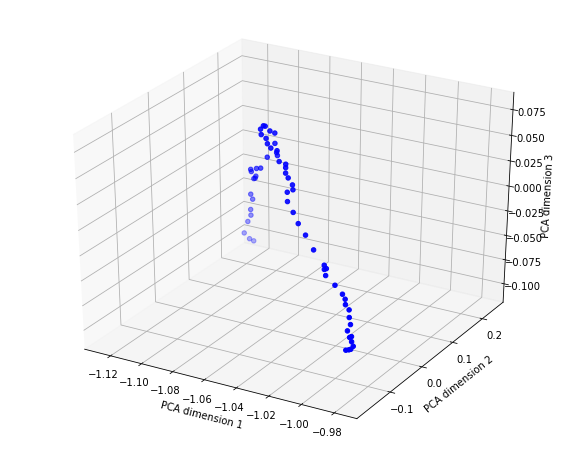}}
  \vspace{-0.3cm}
\end{minipage}
\hfill
\begin{minipage}[b]{0.48\linewidth}
  \centering
  \centerline{\includegraphics[width=4.0cm]{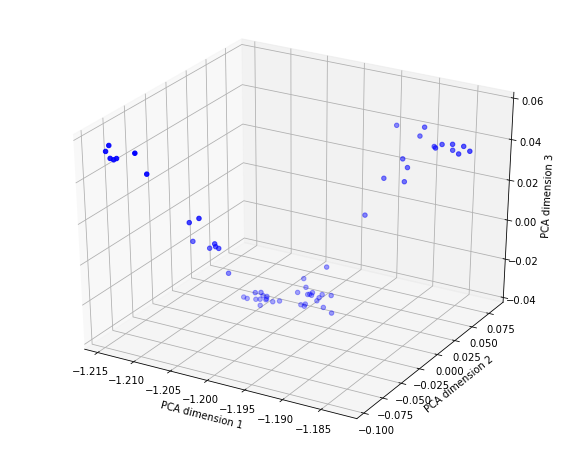}}
  \vspace{-0.3cm}
\end{minipage}
\caption{Neutrophil motion using the first three dimensions of the subspace of the AR model for normal (left) and inhibitor-treated (right). This motion is governed by the AR coefficients.}
\vspace{-0.5cm}
\label{fig:res1}
\end{figure}
\squeezeup
\vspace{-0.20cm}
\subsection{Synthesis}
\label{ssec:submethod3}
\squeezeup
Synthesized neutrophil behavior consists of two parts: content and appearance is sampled from our trained generator $G$ and motion is sampled from a point in subspace $C$. Using \ref{eqn4}, we iteratively generate different sequences. These new sequences are then projected back into the original space, leading to a new motion pattern synthesized entirely from the eigenvector information. The separation of motion and appearance in two streams enable GANs and AR process to identify the respective key features. This results in movement of only the foreground cells and leaves the rest untouched. It also gives an advantage of synthesizing video clips of different cells following different trajectories but nonetheless looks similar to the existing motion patterns. \par
\squeezeup
\vspace{-0.20cm}
\begin{figure}[htb]
\begin{minipage}[b]{1.0\linewidth}
  \centering
  \centerline{\includegraphics[width=8.8cm]{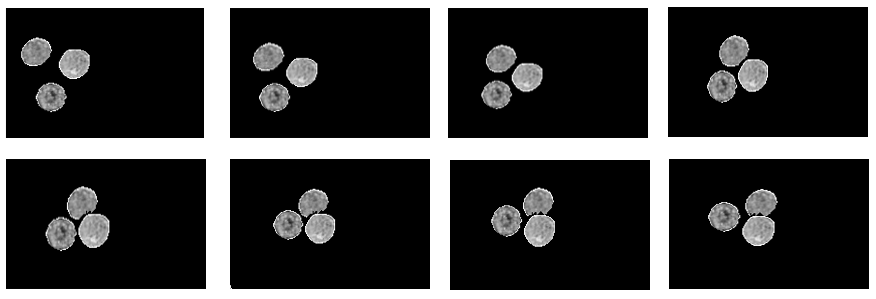}}
  \vspace{-0.3cm}
\end{minipage}
\caption{Sample results of appearance and motion synthesis.}
\vspace{-0.4cm}
\end{figure}
\squeezeup
\squeezeup
\vspace{-3.5mm}
\section{CONCLUSION}
\label{sec:conclusion}
\squeezeup
\vspace{-2.0mm}
In this paper we presented a two stream approach to simulate human neutrophil behavior. Owing to the very limited data at our disposal, we utilized GANs to learn the spatial statistics and AR models to learn the temporal statistics. Bifurcation of appearance and motion allows a controlled video generation process. This work can enable us to quantify changes in organellar appearance, spatial distribution and help in understanding how subsets of the organellar ensembles evolve, improving our understanding of cellular mechanisms as they respond to their environments. 
\squeezeup
\vspace{-4.2mm}
\vspace{-0.1cm}
\section{ACKNOWLEDGMENTS}
\label{sec:ack}
\squeezeup
\vspace{-2.0mm}
We thank R. Ceren for 
constructive criticism of this manuscr-\\-ipt. This work was supported in part by AWS in Education Grant Award. We gratefully acknowledge the support of NVIDIA Corporation with the donation of the Titan X Pascal GPU used for this research.\par
\squeezeup
\vspace{-0.071cm}


\squeezeup
\vspace{-0.05cm}
\bibliographystyle{IEEEbib}
\bibliography{strings,refs}
\end{document}